\DocumentMetadata{lang=en, pdfstandard=ua-2, tagging=on} % enable tagging
\tagpdfsetup{table/header-rows={1}}

\documentclass[11pt]{article}

\usepackage[preprint]{acl}

\usepackage{times}
\usepackage{latexsym}

\usepackage[T1]{fontenc}
\usepackage[utf8]{inputenc}

\usepackage{microtype}

\usepackage{inconsolata}

\usepackage{multirow}
\usepackage{booktabs}
\usepackage[dvipsnames]{xcolor}
\usepackage[most]{tcolorbox}
\usepackage{listings}
\usepackage{csquotes}
\usepackage{subcaption}
\usepackage{pgfplots}
\usepackage{pgfplotstable}
\usetikzlibrary{pgfplots.statistics}
\pgfplotsset{compat=1.18}
\usepackage[section]{placeins}

\title{How Reliable is AI-Generated Audio Description? A Multilingual Expert Evaluation}

\author{Annette Rios \and Lukas Fischer \and Sarah Ebling \\
        Institute of Computational Linguistics \\ University of Zurich \\ \href{mailto:rios@cl.uzh.ch}{rios@cl.uzh.ch}, \href{mailto:lukas.fischer3@uzh.ch}{lukas.fischer3@uzh.ch}, \href{mailto:ebling@cl.uzh.ch}{ebling@cl.uzh.ch}}

\begin{document}
\SuspendTagging{maketitle}
\maketitle
\ResumeTagging{maketitle}
\begin{abstract}
Audio description (AD) makes visual media, such as video, accessible to blind or low-vision (BLV) audiences, but the high cost of professional production limits coverage. Due to recent advances in AI, numerous approaches to automatically generate intermediate AD scripts and/or final AD audios have been proposed. We present an expert evaluation of AI-generated AD scripts for movies in Italian and German that compares two AI systems, a GPT-4o- and a Qwen-based pipeline, against human-authored descriptions. Professional audio describers rated description quality and annotated errors using a detailed error taxonomy that includes content, character, fluency and coherence errors. The results show that while GPT-4o outperforms the Qwen models, both AI systems still fall short of human professional quality: the generated descriptions show substantially higher error rates and lower quality ratings. Based on the error annotations, we identify two key areas that require improvement: 1) character references, which includes both issues with identification and coherent naming, and 2) content errors and hallucinations, where the AD misrepresents or invents events. These error types are particularly critical, as they directly affect the viewer's ability to follow the narrative.
\end{abstract}

\section{Introduction}
Audio description (AD) provides verbal descriptions of important visual elements in videos, films, or live performances to make them accessible to blind or low-vision (BLV) audiences. Secondary target groups include older adults, children or persons with migration background, ADHD, autism spectrum disorder, or viewers who prefer to receive information acoustically or are unable to access visuals for another reason \cite{mazurAudioDescriptionConcepts2020}. The task can be defined as an intermodal, or intersemiotic, translation of visual content into text in the form of a script \cite{Jakobson1959,Geerinck2020}, and ultimately into speech.

Traditionally, audio descriptions are produced by human professionals (often with a background in translation) through a complex, multi-step process. The human audio describer must 1) identify appropriate time slots in which to insert the AD without disrupting the flow of the video and 2) translate visually relevant content into concise descriptions that fit within those temporal constraints. The final step involves recording the script in a studio, ideally with involvement of a blind prooflistener, or synthesizing the speech and mixing the resulting AD audio with the original broadcast soundtrack \cite{gao-etal-2025-audio}. This process is time-consuming and costly, which limits the availability of audio description.
Recent advances in LLMs \cite{brown2020language,touvron2023llama,openai2024gpt4technicalreport} and VLMs \cite{radford2021learning,ramesh2022hierarchical,BLIP} have made the (semi-)automatic generation of audio description, at the intersection of computer vision (CV) and natural language processing (NLP), an active research area \cite{gao-etal-2025-audio}.

However, while generative AI can automate key steps in the generation of AD, current state-of-the-art systems are not yet of sufficient quality to support a fully automated approach. This paper presents an in-depth evaluation and analysis of a state-of-the-art approach to AD generation, where human professional audio describers annotated audio descriptions with error categories and quality ratings in German and Italian. Our analysis reveals systematic differences in AI performance between languages and error types, providing insights that can inform future research on automated AD generation and highlight areas where human expertise remains essential.

\section{Related Work}
As language and video understanding capabilities of current AI models have advanced significantly in recent years, numerous approaches to automatically generating AD have been proposed, with the aim to increase coverage, given the high production costs of expert-authored descriptions \cite{shot-by-shot,Wang2021,NarrAD,AutoAD3,Lee2025amazon,Chu2024,CineAD}. In addition to fully automatic AD generation, collaborative approaches have emerged, where an AI generates initial descriptions that human experts then refine \cite{DescribePro,VideoA11y,ADx3}, based on the observation that fully automated systems still fall short of professional quality standards.

Most of this development in research has been focused on English. Major datasets, such as MAD \cite{MAD} and MADv2 \cite{AutoAD}, LSMDC \cite{LSMDC}, TV-AD \cite{AutoAD-zero}, M-VAD \cite{M-VAD} and CMD-AD \cite{AutoAD3} all contain mostly English data. Likewise, evaluations  and benchmarks for automatic AD generation mostly use English datasets. Furthermore, many of the automatic metrics commonly used for AD evaluation rely on English-specific components, which limits their usage in multilingual settings. CIDEr \cite{Vedantam_2015_CVPR}, for instance, uses the Stanford CoreNLP toolkit for tokenization and stemming. While alternative tokenizers are available in many languages, replacing the preprocessing pipeline can affect the resulting scores, so that a direct comparison with prior work becomes difficult. Additionally, some of the suggested metrics rely on NLP components that are not readily available for other languages, e.g. CRITIC \cite{AutoAD3}, requires coreference resolution, an NLP component with mature tools for English \cite{lee-etal-2017-end,otmazgin-etal-2022-f,joshi-etal-2019-bert,joshi-etal-2020-spanbert} but limited support in most other languages.

User studies and expert evaluations of AI generated descriptions cover some additional languages, e.g., Brazilian Portuguese \cite{Campos2023}, but most studies have been conducted with English-speaking participants \cite{Cheema2025,RNIBreport2025,Stangl2023,Bodi2021}.

Beyond mere coverage, careful evaluations across diverse languages and user groups are also important to reveal language-specific limitations, biases, and errors that can undermine user trust. This is particularly important for applications that provide information to users with sensory disabilities, who cannot independently verify the accuracy of the information provided by AI \cite{Findlater2019,Macleod2017}. As \citeauthor{Bennett2019}~\cite{Bennett2019} point out, AI accessibility solutions that are adopted without critical evaluation risk shifting control away from the users. This is particularly concerning if the system ends up deciding what information the users have access to, with little transparency. Careful assessment of AI capabilities in describing visual content accurately is therefore essential. This paper presents a first evaluation study of AI-generated AD in German and Italian, where professional audio describers annotated descriptions with error categories and quality ratings. While the study is limited to two languages and one annotator per language, our analysis provides initial insights into systematic differences between AI and human-authored AD, and highlights areas where further research and human expertise remain essential.

%%%%%%%%%%%%%%%%%%%%%%%%%%%%%%%%%%%%%%%%%%%%%%%%%%%%%%%%%%%%%%%%%%%%%%%%%%%
\section{Evaluation Design}

\subsection{Data}
We conducted the evaluation on five movies in German and Italian, for which the ground-truth AD was done by SWISS TXT\footnote{SWISS TXT is a subsidiary of the Swiss broadcasting corporation, SRG SSR, that produces audio descriptions for the Swiss market.}, see Table~\ref{tab:movies} in Appendix \ref{app:movies} for a list of the movies. We automatically generated two sets of AD for each movie, using two different models (described in detail in Section \ref{sec:models}). 
Annotations were performed on groups of three consecutive AD segments, as some errors, such as repetitions, can only be assessed in context. For the first AD segment in each set of three segments, we provided the previous description as context, but no video.
For each language, we sampled 80 groups of three consecutive AD segments from each source: two AI systems and a professional audio describer (the original AD). All samples are aligned across sources to cover the same clips, with 10 groups per source repeated for intra-annotator agreement assessment, which results in 720 samples total per language.%\footnote{The dataset (video clips, original and generated AD segments, and annotations will be released upon acceptance.}

\subsection{Evaluated Models}\label{sec:models}

We used the ``Shot-by-Shot'' system \cite{shot-by-shot} to automatically generate audio descriptions for our evaluation. Since we worked with movies in German and Italian instead of English, some adaptations in terms of style (e.g., when to use names vs.\ physical descriptions) and language were necessary. The system is essentially a pipeline that leverages shots\footnote{A shot is a continuous sequence of frames by a single camera, whereas a scene is a collection of shots that narrate a single event or action, usually in one location and timeframe. For instance, a dialogue scene between two characters may consist of alternating shots of each speaker as they talk.} as the fundamental units of video understanding.

The system follows a two-stage framework: a video LLM generates dense descriptions in Stage 1, condensed into AD-style text by a general-purpose LLM in Stage 2, within a word limit derived from the available time slot (we used ground-truth start and end times to enable direct comparison with human-authored AD). Stage 1 draws on preprocessed information that includes shot-level context, narrative threads, character recognition (matching faces to IMDB\footnote{Internet Movie Database:  \url{https://www.imdb.com/}} actor headshots), and shot-scale classification to guide description focus (close up vs. long shots).

We generated two versions of AD for all movies: with an open-source pipeline (Qwen2.5-VL and Qwen3-14B) and a commercial pipeline (GPT-4o for both stages).\footnote{Stage 1: \url{https://huggingface.co/Qwen/Qwen2.5-VL-7B-Instruct}, Stage 2: \url{https://huggingface.co/Qwen/Qwen3-14B}.} Stage 2 prompts were adapted for German and Italian, including language-specific AD guidelines. Stage 1 was kept in English, as preliminary experiments showed that Qwen2.5-VL produces noticeably worse descriptions in German and Italian, with more grammatical errors, less informative content, and lower fluency. The same approach was applied to GPT-4o for consistency.\footnote{The code for the adapted pipeline is available at \url{https://github.com/ZurichNLP/shot-by-shot}. Scripts used to preprocess and sample the data can be found at \url{https://github.com/a-rios/ad-generation}.}

\section{Automatic Quality Assessment}
We find that the Qwen pipeline tends to overgenerate relative to the available time slot, whereas GPT-4o produces descriptions closer to the instructed length. AI-generated AD also tends to be more vague and underspecified than human-authored AD, as indicated by lower named entity counts and higher use of generic words such as {\em jemand/qualcuno} ('someone') or {\em Ding/cosa} ('thing'). For Italian, both AI systems also underuse null subjects compared to human describers, which may be perceived as unnatural by native speakers. ROUGE-L scores \cite{lin-2004-rouge} and full descriptive statistics are provided in Table~\ref{tab:properties}. 
See Appendix~\ref{app:metrics} for details on the automatic quality assessment.

\begin{table*}
 \caption{Properties of Generated and Human AD}\label{tab:properties}
    \begin{subtable}{0.48\linewidth}
    \centering
    \caption{Italian}
    \small
    \begin{tabular}{lrrr}
    \toprule
             & Qwen & GPT-4o  & human\\ \midrule
       ROUGE-L  & 10.00  & 13.67  & -\\
       length ratio AI/human  & 1.48 & 0.95 & -\\
       NER/100 words & 2.80 & 8.27 & 6.56\\
       pronouns/100 words & 11.92 & 7.35 & 7.70\\
       generic words/100 words & 4.30 & 2.14 & 0.28\\
       pro-drop rate & 51.13 & 80.30 & 92.47\\
       \bottomrule
    \end{tabular}
    \end{subtable}\hfill
    \begin{subtable}{0.48\linewidth}
    \caption{German}
    \small
    \begin{tabular}{lrrr}
    \toprule
                & Qwen & GPT-4o  & human\\ \midrule
       ROUGE-L  & 8.19 & 10.41 & - \\
       length ratio AI/human  & 1.42 & 0.80 & - \\
       NER/100 words & 5.94 & 8.94 & 11.01\\
       pronouns/100 words & 10.01 & 10.72 & 9.61 \\
       generic words/100 words & 6.68 & 2.85 & 0.67 \\
       \bottomrule
    \end{tabular}       
    \end{subtable}
\end{table*}

\subsection{Evaluation Interface}

We used Label Studio \cite{Label-Studio} as the annotation interface. Annotators were presented with three consecutive AD segments alongside the corresponding movie clip and asked to assign error labels (multiple-choice with optional free-text) and an overall quality rating (1–5 stars). The quality rating was included as a simple error count may not accurately reflect overall quality, since some errors affect comprehension more than others.
The annotations were carried out by two professional audio describers from SWISS TXT, one specialist for German and one specialist for Italian, both had no prior involvement with the evaluated productions. Samples from all three sources were randomly shuffled; annotators were unaware of the source of each description, and we did not disclose that the original AD was included. Annotators received a short introduction to the interface and detailed guidelines with examples, as well as character name lists with actor headshots to facilitate the annotation of naming errors. Figure~\ref{fig:label-studio} in Appendix \ref{app:interface} shows a screenshot of the interface.

\subsection{Error Classification}
We used four broad categories for the error annotation:
\begin{enumerate}
    \item \textbf{Content:} Errors that relate to the content of the description, e.g. whether it contains redundant or wrong information, or vice versa, omits crucial information.
    \item \textbf{Grammar:} Errors that concern grammar, style or fluency. 
    \item \textbf{Coherence:} Errors where a description does not connect well to the context, e.g., repetitions.
    \item \textbf{Characters:} Errors in character references, e.g., the use of a pronoun for a character that has not yet been introduced by name.
\end{enumerate}

Each of these categories contained a number of more specific error tags for the annotators to choose. Appendix~\ref{app:categories} provides the full list with more details and examples. Overall, the scheme consisted of 24 error categories.

\subsection{Annotation Consistency}

Intra-annotator agreement for overall quality ratings was substantial with Cohen's $\kappa$=0.72 for Italian and $\kappa$=0.80 for German. Note that the Italian annotator did not use the 5-star rating for any description, effectively converting the scale to 4 points. For error tags, agreement was moderate (Jaccard similarity: 0.53$\pm$0.38 Italian, 0.56$\pm$0.44 German), with substantial variation across categories. Objective categories showed higher consistency, such as {\em ungrammatical/not fluent} ($\kappa$=0.56–0.72) and {\em English wording/phrasing} ($\kappa$=0.66–0.74), while more subjective judgments such as {\em contextual gap} ($\kappa$=0.04) and {\em bad character description} ($\kappa$=0.29) were more challenging. Detailed per-category agreement scores are provided in Appendix~\ref{app:agreement}.

%%%%%%%%%%%%%%%%%%%%%%%%%%%%%%%%%%%%%%%%%%%%%%%%%%%%%%%%%%%%%%%%%%%%%%%%%%%
\section{Results}

\subsection{Quality and Error Rates across Sources}

\begin{figure*}
  \centering
  \includegraphics[width=\textwidth, alt={The stacked bar charts show the distribution of quality ratings for each source: Human, GPT-4o and Qwen. The bars show the number of descriptions that received each star rating. Both languages show a clear difference between the quality ratings for human-written AD and automatically generated AD: in the human AD, the vast majority received one of the top two quality ratings, whereas for the generated AD, more than half of the descriptions were ranked as one of the two lowest ratings. This effect is more pronounced with the Qwen models compared to GPT-4o. Table 6 in Appendix C contains the data for this figure.}]{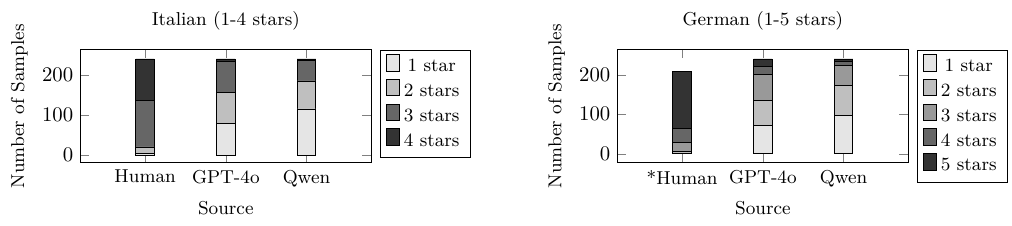}
  \caption{Stars Assigned per Source (*32 German samples containing multilingual dialogue translations were excluded from all results.)}\label{fig:barchart_stars}
\end{figure*}

Human-written AD exhibits substantially fewer errors and higher average ratings than AI-generated descriptions across both languages (Table~\ref{tab:error_rates}). GPT-4o achieves lower error rates and higher quality ratings than the Qwen pipeline; this is expected, given the considerably smaller model sizes (7B/14B parameters). The Italian annotator never assigned the maximum 5-star rating, even for human-authored descriptions, which makes the rating for Italian effectively a 4-point scale. There is a stark contrast between human-authored descriptions and their AI-generated counterparts in terms of quality assessment, as shown in Fig.~\ref{fig:barchart_stars}.\footnote{Note that 32 German samples from one movie contained translations of multilingual dialogue in the original (human) script, rather than audio descriptions. The annotator rated these as 1 star with the explanation that they are not proper audio descriptions (but they are not poor quality). To keep the evaluation strictly on audio description quality, we excluded these 32 samples from all results presented in the following sections.}

\begin{table*}
    \centering
    \caption{Error Rates and Average Rating per Source}\label{tab:error_rates}
    \begin{subtable}{0.48\linewidth}
    \caption{Italian}
        \begin{tabular}{lccc}
    \toprule
        Source &  total errors & error rate & $\bar{x}$ stars (1-4)  \\ \midrule
         human & 91 & 0.4 & 3.3 \\
         GPT-4o & 613 & 2.6 & 2.0 \\
         Qwen & 839 & 3.5 & 1.8 \\
         \bottomrule
    \end{tabular}
    \end{subtable}
    \hfill
    \begin{subtable}{0.48\linewidth}
        \caption{German}
         \begin{tabular}{lccc}
    \toprule
        Source  & total errors & error rate & $\bar{x}$ stars (1-5) \\ \midrule
         human & 27 & 0.1 & 4.1\\
         GPT-4o & 260 & 1.1 & 2.4\\
         Qwen & 359 & 1.5 & 2.0\\
         \bottomrule
    \end{tabular}
    \end{subtable}
   
\end{table*}

\subsection{How Errors Affect Quality Ratings}

\begin{figure*}
  \centering
  \includegraphics[width=\textwidth, alt={The figure shows box plots that illustrate the error counts for each star rating. In both languages, there is a clear negative correlation between the number of errors and the quality rating assigned by the annotators. Descriptions that are rated with one star have the highest median error count and also the largest spread, up to a maximum of twelve errors in both Italian and German. Higher rated descriptions have fewer errors and less variance, in both languages, the median error count for the highest rating is zero. The negative correlation is consistent across both languages, but Italian has a considerably higher median error count than German in the lowest rated audio descriptions. Table 7 in Appendix C contains the data for this figure.}]{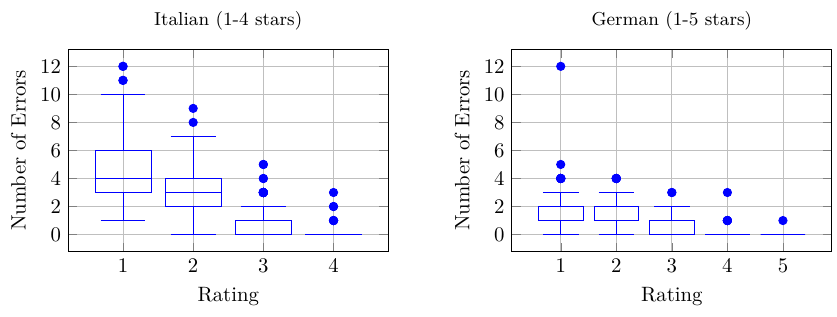}
  \caption{Error Counts and Quality}\label{fig:error_rating_correlation}
\end{figure*}

\begin{figure*}
  \centering
  \includegraphics[width=\textwidth, alt={The figure shows bar charts for the three errors with the highest and the three errors with the lowest impact on quality rating. The bars express the star penalty, i.e. the average rating decrease when the error is present. Additionally, each error is labeled with the Spearman's rho, which indicates the negative correlation of the error with the rating assigned to the description. The error categories with the highest star penalty involve errors in character identification, or errors where the description is factually incorrect or unrelated to the scene. Errors with the lowest star penalty are redundancy errors and the omission of on-screen text in the audio description. Table 8 in Appendix C contains the data for this figure.}]{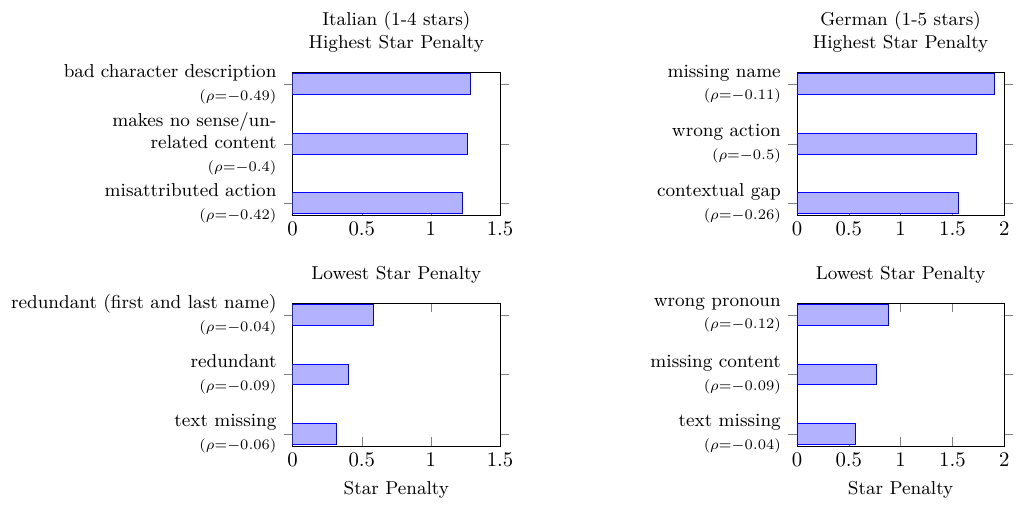}
  \caption{Impact on Rating: Star Penalty and Spearman's $\rho$ for Tags n$>$5}\label{fig:error_severity}
\end{figure*}

There is a clear cumulative effect of errors: the more error tags were assigned to an audio description, the lower its quality rating, with a near-perfect linear fit (R$^2$=0.964 for Italian, R$^2$=0.923 for German). Figure ~\ref{fig:error_rating_correlation} shows the distribution of error counts per rating.
To understand which specific errors affect quality rating the most, we calculated the practical impact of each error type, quantified as {\em star penalty}, i.e. the average rating decrease when that error is present. Figure~\ref{fig:error_severity} shows the three errors with the highest and the three errors with the lowest star penalty. The Spearman's $\rho$ values indicate the negative correlation between error presence and rating.

In both languages, errors with the highest impact on quality ratings involve either character identification ({\em missing name}, {\em bad character description}) or content misrepresentation ({\em wrong action}, {\em unrelated content}, {\em misattributed action}). The error with the lowest impact in both languages is {\em text missing}, i.e., on-screen text not included in the AD. The severity of this error depends on how relevant the text is to the plot.

%%%%%%%%%%%%%%%%%%%%%%%%%%%%%%%%%%%%%%%%%%%%%%%%%%%%%%%%%%%%%%%%%%%%%%%%%%%
\subsection{Error Composition}
The Italian AD, both human and AI-generated, contains substantially more errors than the German AD. Since we have only one annotator per language, we cannot reliably distinguish whether this reflects genuine quality differences or annotator strictness. However, language-related errors such as {\em ungrammatical/not fluent} and {\em English wording/phrasing (Itanglese/Denglisch)} are relatively objective and show a larger cross-lingual gap than other error types, which suggests that the models genuinely struggled more with Italian. Notably, German outputs showed no English interference, whereas both models received this label in Italian.
In both languages, the few errors in human-written AD were mostly grammatical or fluency mistakes. In Italian, human describers also showed a higher rate of omitted character names. This might reflect an annotation artifact, as without access to the full movie context, name omissions (e.g. null subjects) that would be unambiguous to a viewer may appear ambiguous to the annotator, since they only see the preceding 1-3 descriptions of the corresponding clip. The {\em makes no sense/unrelated content} category was exclusive to AI-generated AD in both languages.
Error composition details are shown in heatmaps in Appendix~\ref{app:heatmaps}.

%%%%%%%%%%%%%%%%%%%%%%%%%%%%%%%%%%%%%%%%%%%%%%%%%%%%%%%%%%%%%%%%%%%%%%%%%%%
\section{Conclusions}
This paper presents an expert evaluation of AI-generated audio description scripts in Italian and German, two languages that have not received much attention so far in research on automatic AD generation. The results show that the AI-generated audio descriptions are still well below the quality of professional, human-authored descriptions. When human describers make errors, they mostly make grammatical or fluency mistakes rather than content errors. AI-generated descriptions, on the other hand, show substantially higher rates of errors that directly affect narrative comprehension, such as hallucinated content, character identification failures and wrong descriptions of objects or actions.

Among the evaluated systems, GPT-4o consistently outperforms the Qwen pipeline across both languages and most error categories. The gap is particularly notable in Italian, where both models showed higher rates of language-related errors compared to German. We note, however, that the field is rapidly evolving. The models evaluated here, which represented the latest available versions at the time of data generation for this study, have already been replaced by newer releases. 

Based on our findings, there are two key areas that require improvement for AI-generated AD: 
\begin{description} 
   \item[Character references:] Character identification remains challenging, as factors such as lighting, camera angle, or partial occlusion can make identification difficult. Furthermore, consistency in character naming (when to use a pronoun and when to use the name) and character descriptions remain challenging, and such inconsistencies can make it difficult for viewers to follow the narrative.
    \item[Content errors and hallucinations:] Errors where the description does not match what is happening on screen are among the most common issues in both languages and can completely disrupt a viewer's understanding of the plot.
\end{description}

As AI-generated AD becomes more widely deployed, careful and ongoing expert evaluation across diverse languages remains essential to ensure that the quality of the generated descriptions meets the standards that blind and low-vision viewers depend on.

This study provides a first step in that direction, though it is limited to two languages and a single annotator per language. This setting limits the cross-lingual comparison, as observed differences between Italian and German may in part reflect annotator differences rather than a genuine variation in model performance. 
Additionally, our evaluation covers only two European languages and a single genre, movies. Evaluations on other formats, such as documentaries, news or sports broadcasts, might require an adaption of the error categories and could potentially exhibit different error profiles.

Future work should involve multiple annotators and extend to other languages and genres.

\section*{Acknowledgments}
This work was funded by the Swiss Innovation Agency (Innosuisse) Flagship Inclusive Information and Communication Technologies (IICT) under grant agreement PFFS-21-47. We also thank our industry partners SWISS TXT and SRG, particularly, Veronica Leoni and Fabio Ben Soltane, for their time and expertise in providing the error annotations. We are also grateful to Alexa Lintner for feedback on the annotation interface and error categories.

% % %%%%%%%%%%%%%%%%%%%%%%%%%%%%%%%%%%%%%%%%%%%%%%%%%%%%%%%%%%
% % % Custom bibliography entries only
\SuspendTagging{biblio}
 \bibliography{bib/accessibility,bib/nlp}
\ResumeTagging{biblio} 
% %%%%%%%%%%%%%%%%%%%%%%%%%%%%%%%%%%%%%%%%%%%%%%%%%%%%%%%%%%

\appendix
\onecolumn
\section{Evaluation Design}\label{sec:appendix}

\subsection{Evaluated Movies}\label{app:movies}
\begin{table}[ht]
 \caption{German and Italian movies used in the evaluation. IMDB title in brackets, if applicable.}
    \label{tab:movies}

    \begin{tabular}{llrrr}
    \toprule
         &  Title & year &\#AD & IMDB ID\\ \midrule \addlinespace
   \multirow{4}{1.3cm}{German} & Baghdad in my Shadow & 2019 & 729 & tt6864088 \\
         & Moskau Einfach ({\em One-way to Moscow}) & 2020& 618 & tt10949014 \\
         & Trommelwirbel ({\em Tambour battant})& 2019 &418 & tt9558602\\
         & 8 Tage im August & 2023 & 1039 & tt28132461\\ \addlinespace \cmidrule{2-5} \addlinespace

    \multirow{4}{1.3cm}{Italian} & La Tentazione di Esistere & 2022 & 519 & tt24248964  \\
         & Quanto Basta ({\em As needed}) & 2018 & 419 & tt7117552 \\
         & Riders of Justice & 2020 & 524 & tt11655202\\
         & Stürm ({\em Caged Birds})& 2020 & 631 & tt9502538 \\ \bottomrule   
    \end{tabular}
\end{table}

\subsection{Statistics of Generated and Human AD}\label{app:metrics}
Table~\ref{tab:properties} reports ROUGE-L\footnote{ROUGE-L measures the longest common subsequence overlap between generated and reference text \cite{lin-2004-rouge}.} scores and descriptive statistics for both systems. Named entity density was computed using spaCy\footnote{Italian: it\_core\_news\_lg, German: de\_core\_news\_lg}. The generic word frequency is based on a manually compiled list of generic nouns and pronouns in both languages: 
\begin{description}
    \item[Italian]: {\em cosa, oggetto, elemento, persona, qualcuno, qualcosa, posto, luogo, gente, uomo, donna, chiunque, tipo, individuo, figura, soggetto, ambiente, zona, spazio, gesto, gestire, gesticolare}
    \item[German]: {\em Ding, Sache, Gegenstand, Objekt, Person, jemand, Etwas, Ort, Stelle, Bereich, Leute, Person, Mann, Frau, Typ, Kerl, Mensch, Raum, Bereich, Stelle, Gegend, Geste, gestikulieren}
\end{description}

For Italian, we additionally report the pro-drop rate (null subjects per 100 finite verbs with a pronominal subject). Human-authored AD drop the majority of pronominal subjects, whereas both AI systems tended to use overt pronouns or character names instead (this might in part explain the high NER density for GPT-4o in Italian). See the following example, where (1) is the description produced by GPT-4o and (2) is the human description:

\begin{quote}% tabular, gb4e breaks tagging with pdflatex
(1)\\[2pt]
\begin{tabular}{@{}lllllllllll@{}}
\textbf{Arturo} & si & siede & di & fronte & a & \textbf{Corradi} & al & tavolo & con un & bicchiere\\
Arturo & rflx & sit.down & in & front & of & Corradi & at & table & with a & glass \\
\end{tabular}\\[2pt]
\enquote{Arturo sits down opposite Corradi at the table with a glass.}

\medskip

(2)\\[2pt]
\begin{tabular}{@{}llllllll@{}}
Si & siedono & a & un & tavolo & uno di & fronte & all'altro.\\
rflx & $\emptyset$-sit.down & at & a & table & one & in front & of.other\\
\end{tabular}\\[2pt]
\enquote{They sit down at a table facing each other.}
\end{quote}

\clearpage
\subsection{Annotation Interface}\label{app:interface}
\nopagebreak[4]
\begin{figure}[!h]
    \centering
    \includegraphics[width=0.9\linewidth, alt={A screenshot of the annotation interface, that shows a video clip on top, with two men walking through a park with a small, white dog on the leash. Below the video, the first of three audio descriptions is shown in text, with a multiple choice form where annotators can select specific error tags. Below that form, there is a quality rating where annotators can select between one and five stars, and an optional text field where annotators can leave a comment.}]{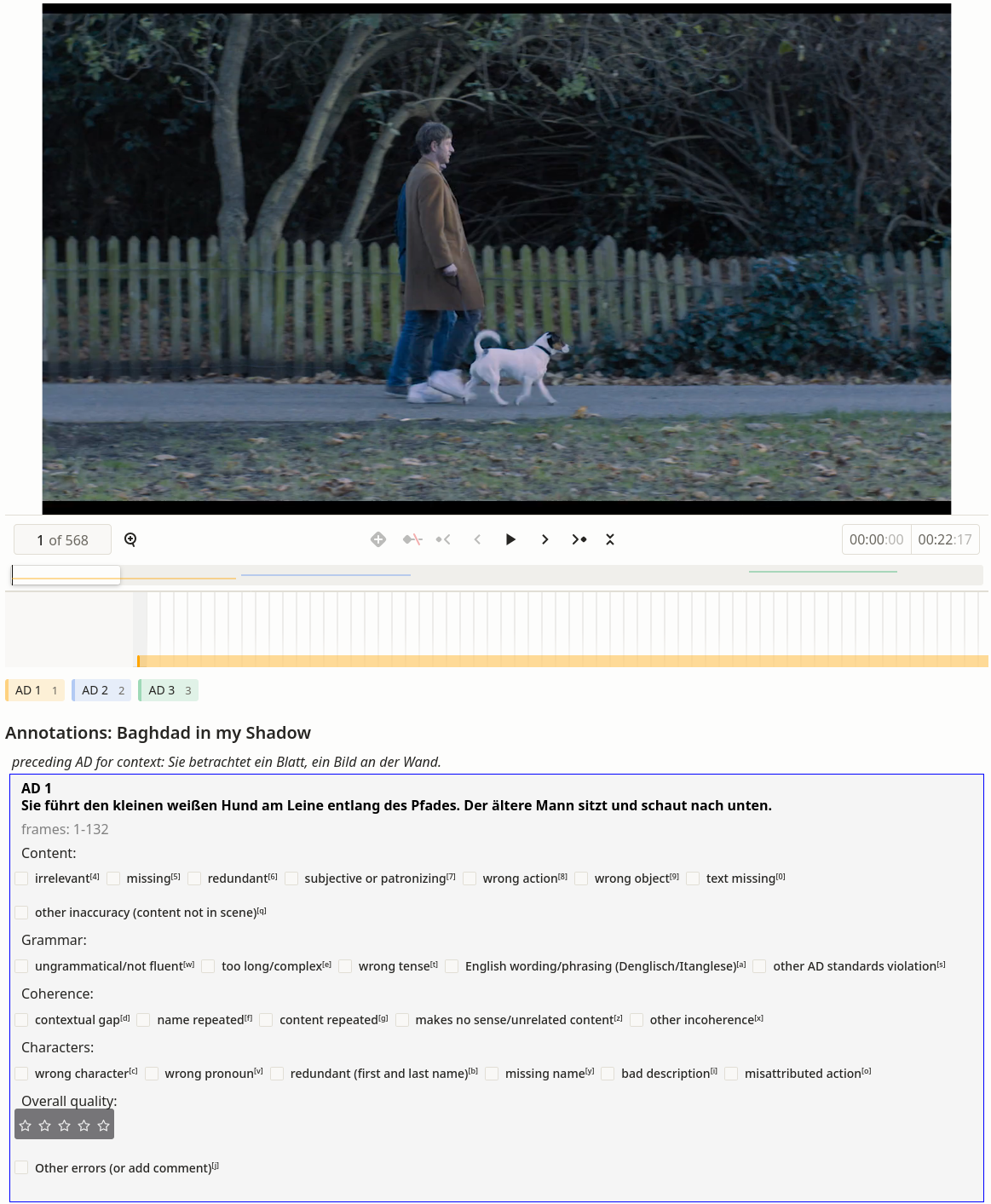}
    \caption{Annotation interface: Three consecutive AD segments are given, with the corresponding movie clip. Only the first of three AD segments is shown here, due to space.}
    %\Description{A screenshot of the annotation interface, that shows a video clip on top, with two men walking through a park with a small, white dog on the leash. Below the video, the first of three audio descriptions is shown in text, with a multiple choice form where annotators can select specific error tags. Below that form, there is a quality rating where annotators can select between one and five stars, and an optional text field where annotators can leave a comment.}
    \label{fig:label-studio}
\end{figure}
\clearpage
%%%%%%%%%%%%%%%%%%%%%%%%%%%%%%%%%%%%%%%%%%%%%%%%%%%%%%%%%%%%%%%%%%%%%%%
\subsection{Error Categories}\label{app:categories}
Since we use data provided by SWISS TXT in German and Italian, we follow their guidelines for the definition of error categories. While most error types are problematic across audio descriptions generally, some categories may be specific to our dataset and languages.

\begin{table}
\small
\caption{Error taxonomy: all tags with descriptions and examples.}\label{tab:taxonomy}
\begin{tabular}{p{0.09\textwidth}p{0.19\textwidth}p{0.62\textwidth}}
\toprule
Category & Tag & Description and examples \\ \midrule
Content & Irrelevant & AD contains information that is not relevant for the viewer, e.g.\ a description of a background object that is not needed for the storytelling. \\
Content & Missing content & Crucial information needed to follow the plot is missing, or the AD is too vague to be useful, e.g.\ \textit{she holds an object}. \\
Content & Redundant & AD repeats information that is obvious from the audio, e.g.\ {\em the phone is ringing}. \\
Content & Subjective or patronizing & AD is not neutral, e.g.\ {\em the kitchen is a mess} instead of {\em dirty dishes are stacked on the counter}. \\
Content & Wrong action & AD describes an action incorrectly, e.g.\ {\em a man walks down the street}, but in the video the character is running. \\
Content & Wrong object & An object was not correctly identified, e.g.\ {\em he holds a book}, but in the video the character is holding a phone. \\
Content & Text missing & Relevant on-screen text is not included in the AD, e.g.\ a note that is shown on screen, credits, or text overlays such as {\em 5 years earlier} introducing a flashback. \\
Content & Other inaccuracy & Wrong information of another type that is not covered by the given categories. \\
Fluency & Ungrammatical/not fluent & AD contains grammatical errors, typos, or awkward phrasing. \\
Fluency & Too long/complex & AD is too long and/or complex, e.g.\ contains nested sentences. \\
Fluency & Wrong tense & AD is not in present tense, e.g.\ {\em Lisa \textbf{sat} down} instead of {\em Lisa sits down}. \\
Fluency & English wording/phrasing & AD contains words or phrases that are literal translations from English, e.g.\ German {\em Sie betrachtet sich im Spiegel und \textbf{justiert} ihr Haar} (\enquote{she looks at herself in the mirror and adjusts her hair}), or Italian {\em Lei \textbf{si rompe giù} in lacrime} (\enquote{she breaks down in tears}). \\
Fluency & Other AD standards violation & The description violates AD standards or guidelines in a way that is not covered by the other tags. \\
Coherence & Contextual gap & An information gap to the preceding description leads to confusion, e.g.\ AD~1: {\em Paul and Lisa are talking to each other in front of the car.} AD~2: {\em Paul closes the door and leans back.} (Paul getting into the car was not mentioned.) \\
Coherence & Name repeated & The same name is repeated from the previous segment instead of using a pronoun, e.g.\ AD~1: {\em \textbf{Paul} sits down at the table.} AD~2: {\em \textbf{Paul} starts eating.} \\
Coherence & Content repeated & Content is repeated from the previous segment, e.g.\ AD~1: {\em Paul is talking on the phone, walking down the street.} AD~2: {\em Paul talks on the phone.} \\
Coherence & Makes no sense/unrelated content & AD makes no sense with respect to the clip and/or has completely unrelated content. \\
Coherence & Other incoherence & AD is not properly \enquote{connected}, but none of the other categories match. \\
Character & Wrong character & AD contains a character name that does not appear in the clip. \\
Character & Wrong pronoun & AD uses an incorrect pronoun. We observed some bias, e.g.\ models sometimes use female pronouns for male characters with long hair. \\
Character & Redundant name & AD contains the correct name, but uses first and last name where the last name is not needed and sounds unnatural. \\
Character & Missing name & AD refers to an already introduced character with a description, or uses a pronoun where the name is needed to avoid confusion with another character. \\
Character & Bad character description & AD describes a character by physical attributes that are wrong ({\em a blonde man}, but the person has dark hair), inappropriate, or too vague to be useful ({\em a person in a shirt}). \\
Character & Misattributed action & Characters are recognized correctly, but subject and/or object of the action are confused, e.g.\ the scene is {\em A looks at B}, but the AD says {\em B looks at A}. \\
\bottomrule
\end{tabular}
\end{table}
%%%%%%%%%%%%%%%%%%%%%%%%%%%%%%%%%%%%%%%%%%%%%%%%%%%%%%%%%%%%%%%%%%%%%%%
\clearpage
\subsection{Intra-annotator Agreement}\label{app:agreement}
For the 24 error categories, we used Jaccard similarity to assess consistency, since annotators could assign multiple tags per sample. Given the limited size of our repeated samples (30 video clips, 90 descriptions), many error categories did not have sufficient instances for reliable agreement estimation. Table~\ref{tab:agreement} presents Cohen's $\kappa$\footnote{Each tag is treated as a binary label (present/absent) per sample for calculating Cohen's $\kappa$.},precision, recall, and F1 scores for categories with $\geq$15 occurrences. F1, precision and recall are computed by treating the first annotation pass as the reference and the second pass as the predicted labels.

\begin{table}[ht]
    \centering
     \caption{Agreement Scores for Tags with Count$>$15 (F1: Pass 2 vs. Pass 1)}\label{tab:agreement}
     \begin{subtable}{\linewidth}
     \centering
     \caption{Italian}
         \begin{tabular}{lrrrrr}
         \toprule
        Category &  Counts & $\kappa$ & F1 & Precision & Recall \\ \midrule
        ungrammatical/not fluent & 64 & 0.56 & 0.72  & 0.72  & 0.72 \\
        English wording/phrasing (Denglisch/Itanglese) & 43 & 0.66 & 0.74 & 0.70 &  0.80\\
        bad character description & 37 & 0.29 & 0.43  & 0.53 & 0.36 \\
        wrong action & 31 &  0.26 & 0.39 &  0.33 & 0.46  \\ 
        misattributed action & 29 & 0.63 & 0.69 & 0.71 & 0.67\\
        makes no sense/unrelated content & 29 & 0.31 & 0.41 & 0.35 & 0.5\\
        wrong character & 26 & 0.28 & 0.38 & 0.45 & 0.33\\
        missing name  & 26 & 0.19 & 0.31 & 0.29 & 0.33 \\ \bottomrule
        \end{tabular}
     \end{subtable}\bigskip
    
     \begin{subtable}{\linewidth}
     \centering
     \caption{German}
     \begin{tabular}{lrrrrr}
        \toprule
        Category &  Counts & $\kappa$ & F1 & Precision & Recall \\ \midrule
        wrong action & 33 & 0.45 & 0.55 &  0.75 & 0.43 \\
        wrong object & 18 & 0.63 & 0.67 & 0.86 & 0.55 \\
        contextual gap & 16 & 0.04 & 0.13 & 0.13 & 0.13 \\
        wrong character & 16 & 0.32 & 0.38 & 0.30 & 0.50 \\
         \bottomrule
     \end{tabular}  
     \end{subtable} 
\end{table}

%%%%%%%%%%%%%%%%%%%%%%%%%%%%%%%%%%%%%%%%%%%%%%%%%%%%%%%%%%%%%%%%%%%%%%%
\clearpage
\section{Results}
\subsection{Error Compositions}\label{app:heatmaps}
\begin{figure}[h]
 \centering
 \caption{Error Composition as Heatmaps.}\label{fig:heatmaps}
 %\Description{The figure shows heatmaps for both languages that reflect the relative proportion of each error type across the three sources: Qwen, GPT-4o and human. In both languages, the most common error in human-written AD are grammatical and fluency issues. In Italian, the human AD also shows a fair proportion of \enquote{missing name} errors. Grammar and fluency issues are also present in AI-generated AD, however, the overall error distribution is considerably more even across error types compared to the human-written AD. In other words, the AI models produce more errors in absolute terms, and their errors are also more diverse compared to human describers.}
   \begin{subfigure}{0.48\textwidth}
   \caption{Italian}
       \includegraphics[width=\linewidth, alt={A heatmap that shows the relative error distribution for Italian. Rows are error types grouped by category, columns are the three sources (Qwen, GPT-4o, human), and each cell shows that error type's percentage of the source's total errors. In the human descriptions, grammar and fluency errors and missing names stand out as the most frequent. Grammar and fluency errors are also the most common in the AI-generated descriptions, but their errors are spread more evenly across categories. Table 9 a) in Appendix C contains the data for this figure.}]{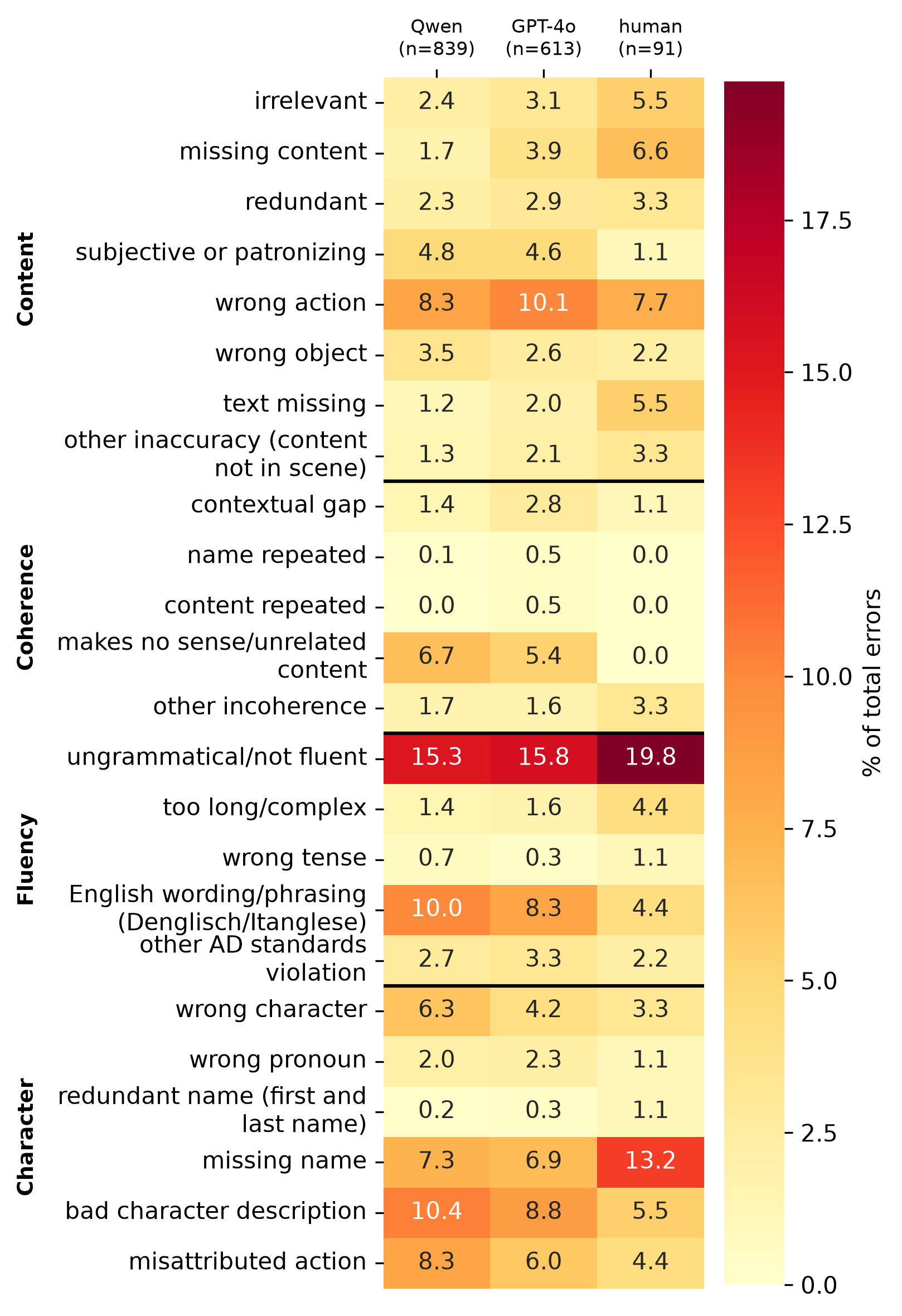}
   \end{subfigure}\hfill
   \begin{subfigure}{0.48\textwidth}
   \caption{German}
       \includegraphics[width=\linewidth, alt={A heatmap that shows the relative error distribution for German. Layout is identical to the Italian panel: rows are error types grouped by category, columns are the three sources (Qwen, GPT-4o, human), and each cell shows that error type's percentage of the source's total errors. In the human descriptions, grammatical and fluency errors stand out as by far the most frequent. These are also common in the AI-generated descriptions, but the single most frequent AI error is wrong action, followed by other content and character errors such as wrong object, wrong character, wrong pronoun and bad character description. Table 9 b) in Appendix C contains the data for this figure.}]{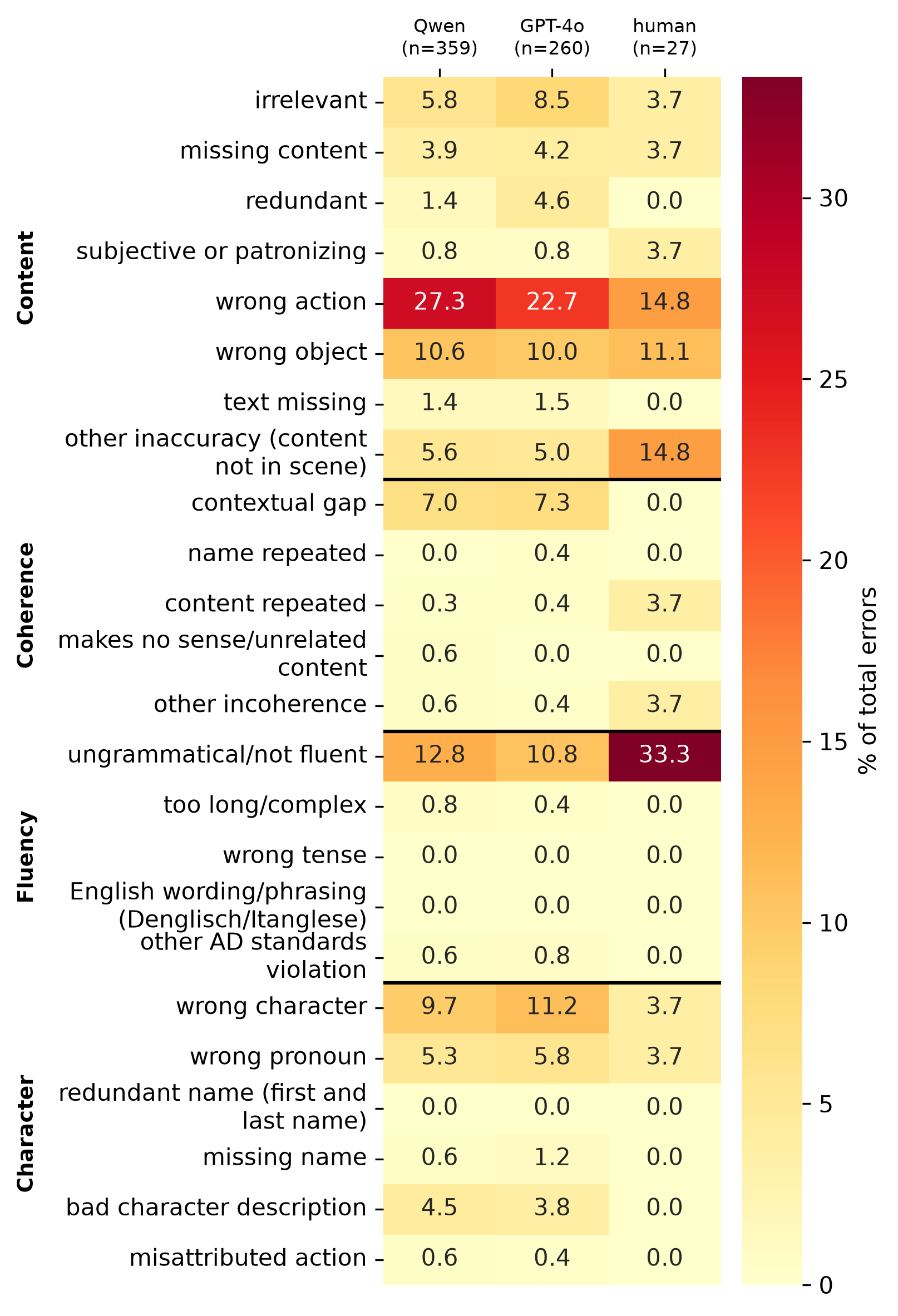}
   \end{subfigure}
\end{figure}

Fig. \ref{fig:heatmaps} shows the error composition as heatmaps for both languages. Note that the Italian annotator used all error categories, whereas the German annotator never assigned the tags {\em wrong tense}, {\em English wording/phrasing (Denglisch/Itanglese)}, {\em redundant name (first and last name)}. 
Another observation on the relatively high proportion of the error
{\em missing name} in Italian: This might reflect an annotation artifact, as without access to the full movie context, name omissions (e.g. null subjects) that would be unambiguous to a viewer may appear ambiguous to the annotator, since they only see the preceding 1-3 descriptions of the corresponding clip.
In both languages, the most common error in human-written AD are grammatical and fluency issues. Grammar and fluency issues are also present in AI-generated AD, however, the overall error distribution is considerably more even across error types compared to the human-written AD. In other words, the AI models produce more errors in absolute terms, and their errors are also more diverse compared to human describers.

%%%%%%%%%%%%%%%%%%%%%%%%%%%%%%%%%%%%%%%%%%%%%%%%%%%%%%%%%%%%%%%%%%%%%%%
\clearpage
\section{Figure Data in Tabular Format}\label{app:figure-as-table}
\begin{table}[ht]
 \centering
     \caption{Stars Assigned per Source. Data for Fig.~\ref{fig:barchart_stars}.}
        \label{tab:barchart_stars}
        \begin{subtable}{0.48\linewidth}
        \centering
        \caption{Italian}
        \small
            \begin{tabular}{crrr}
            \toprule
            Stars & Human & GPT-4o & Qwen \\ \midrule
           1  & 4 & 79 & 115\\
           2  & 16 & 77 & 70\\
           3 & 117 & 78 & 51\\
           4 & 103 & 6 & 4\\
           \bottomrule
        \end{tabular}
        \end{subtable}
        \begin{subtable}{0.48\linewidth}
        \centering
        \caption{German}
        \small
            \begin{tabular}{crrr}
            \toprule
           Stars  & Human & GPT-4o & Qwen \\ \midrule
             1 & 1 & 72 & 96 \\
             2 & 5 & 62 & 76\\
             3 & 22 & 66& 51\\
             4 & 35 & 21 & 10\\
             5 & 145 & 19 & 7\\
             \bottomrule
        \end{tabular}
        \end{subtable}  
\end{table}

\begin{table}[ht]
    \centering
    \caption{Error Counts and Quality. Data for Fig.~\ref{tab:error_rating_correlation}.}\label{tab:error_rating_correlation}
     \begin{subtable}{0.48\linewidth}
     \centering
     \caption{Italian}
     \small
         \begin{tabular}{lrrrr}
         \toprule
                  & 1 Star & 2 Stars & 3 Stars & 4 Stars \\ \midrule
           median  & 4 & 3 & 0 & 0 \\
           Q1  & 3 & 2 & 0 & 0\\
           Q3 & 6 & 4 & 1 & 0 \\
           min & 1 & 0 & 0 & 0 \\
           max & 12 & 9 & 5 & 3\\
           n & 198 & 163 & 246 & 113\\
           \bottomrule
         \end{tabular}
     \end{subtable}
     \begin{subtable}{0.48\linewidth}
     \centering
         \caption{German}
         \small
      \begin{tabular}{lrrrrr}
      \toprule
         & 1 Star & 2 Stars & 3 Stars & 4 Stars & 5 stars \\ \midrule
           median  & 2 & 1 & 1 & 0 & 0\\
           Q1 & 1 & 1 & 0 & 0 & 0\\
           Q3 & 2 & 2 & 1 & 0 & 0\\
           min & 0 & 0 & 0 & 0 & 0\\
           max & 12 & 4 & 3 & 3 & 1\\
           n & 169 & 143 & 139 & 66 & 171\\
           \bottomrule
         \end{tabular}
     \end{subtable} 
\end{table}

\begin{table}[ht]
    \centering
    \caption{Impact on Rating: Star Penalty for Tags $n>5$. Data for Fig.~\ref{fig:error_severity}.}\label{tab:error_severity}
     \begin{subtable}{0.59\linewidth}
     \centering
     \caption{Italian}
         \begin{tabular}{lrr}
         \toprule
         Error & Star Penalty  & n\\ \midrule
         Highest Impact: \\
         bad character description & 1.28 & 146\\
         makes no sense/unrelated content & 1.27  & 89\\
         misattributed action & 1.23  & 111\\ \addlinespace
         Lowest Impact: \\
         redundant (first and last name) & 0.58  & 5 \\
         redundant & 0.40  & 40\\
         text missing & 0.32 & 27\\ 
           \bottomrule
         \end{tabular}
     \end{subtable}\hfill
     \begin{subtable}{0.39\linewidth}
     \centering
         \caption{German}
      \begin{tabular}{lrrrr}
      \toprule
         Error & Star Penalty  & n\\ \midrule
         Highest Impact: \\
         missing name & 1.91  & 5 \\
         wrong action & 1.73  & 161 \\
         contextual gap & 1.56  & 44 \\ \addlinespace
         Lowest Impact: \\
         wrong pronoun & 0.88 & 35 \\
         missing content & 0.77  & 26\\
         text missing & 0.57 & 9\\ 
           \bottomrule
        
         \end{tabular}
     \end{subtable} 
\end{table}

\begin{table}[h]
\centering
 \caption{Data for Error Distribution Heatmaps (Fig.~\ref{fig:heatmaps}): Error Composition per Source, in Percentage of Total Errors.}
 \begin{subtable}{0.9\linewidth}
 \centering
     \caption{Italian}
     \small
        \begin{tabular}{lrrr}
        \toprule
        & Qwen (n=839) & GPT-4o (n=613) & human (n=91) \\
        \midrule
        irrelevant & 2.4 & 3.1 & 5.5 \\
        missing content & 1.7 & 3.9 & 6.6 \\
        redundant & 2.3 & 2.9 & 3.3 \\
        subjective or patronizing & 4.8 & 4.6 & 1.1 \\
        wrong action & 8.3 & 10.1 & 7.7 \\
        wrong object & 3.5 & 2.6 & 2.2 \\
        text missing & 1.2 & 2.0 & 5.5 \\
        other inaccuracy (content not in scene) & 1.3 & 2.1 & 3.3 \\
        contextual gap & 1.4 & 2.8 & 1.1 \\
        name repeated & 0.1 & 0.5 & 0.0 \\
        content repeated & 0.0 & 0.5 & 0.0 \\
        makes no sense/unrelated content & 6.7 & 5.4 & 0.0 \\
        other incoherence & 1.7 & 1.6 & 3.3 \\
        ungrammatical/not fluent & 15.3 & 15.8 & 19.8 \\
        too long/complex & 1.4 & 1.6 & 4.4 \\
        wrong tense & 0.7 & 0.3 & 1.1 \\
        English wording/phrasing (Denglisch/Itanglese) & 10.0 & 8.3 & 4.4 \\
        other AD standards violation & 2.7 & 3.3 & 2.2 \\
        wrong character & 6.3 & 4.2 & 3.3 \\
        wrong pronoun & 2.0 & 2.3 & 1.1 \\
        redundant name (first and last name) & 0.2 & 0.3 & 1.1 \\
        missing name & 7.3 & 6.9 & 13.2 \\
        bad character description & 10.4 & 8.8 & 5.5 \\
        misattributed action & 8.3 & 6.0 & 4.4 \\
        \bottomrule
        \end{tabular}
 \end{subtable}\\ \vspace{5pt}
 \begin{subtable}{0.9\linewidth}
 \centering
     \caption{German}
     \small
        \begin{tabular}{lrrr}
        \toprule
         & Qwen (n=359) & GPT-4o (n=260) & human (n=27) \\
        \midrule
        irrelevant & 5.8 & 8.5 & 3.7 \\
        missing content & 3.9 & 4.2 & 3.7 \\
        redundant & 1.4 & 4.6 & 0.0 \\
        subjective or patronizing & 0.8 & 0.8 & 3.7 \\
        wrong action & 27.3 & 22.7 & 14.8 \\
        wrong object & 10.6 & 10.0 & 11.1 \\
        text missing & 1.4 & 1.5 & 0.0 \\
        other inaccuracy (content not in scene) & 5.6 & 5.0 & 14.8 \\
        contextual gap & 7.0 & 7.3 & 0.0 \\
        name repeated & 0.0 & 0.4 & 0.0 \\
        content repeated & 0.3 & 0.4 & 3.7 \\
        makes no sense/unrelated content & 0.6 & 0.0 & 0.0 \\
        other incoherence & 0.6 & 0.4 & 3.7 \\
        ungrammatical/not fluent & 12.8 & 10.8 & 33.3 \\
        too long/complex & 0.8 & 0.4 & 0.0 \\
        wrong tense & 0.0 & 0.0 & 0.0 \\
        English wording/phrasing (Denglisch/Itanglese) & 0.0 & 0.0 & 0.0 \\
        other AD standards violation & 0.6 & 0.8 & 0.0 \\
        wrong character & 9.7 & 11.2 & 3.7 \\
        wrong pronoun & 5.3 & 5.8 & 3.7 \\
        redundant name (first and last name) & 0.0 & 0.0 & 0.0 \\
        missing name & 0.6 & 1.2 & 0.0 \\
        bad character description & 4.5 & 3.8 & 0.0 \\
        misattributed action & 0.6 & 0.4 & 0.0 \\
        \bottomrule
        \end{tabular}
 \end{subtable}
\end{table}

\end{document}